\documentclass[runningheads]{llncs}
\pdfoutput=1
\usepackage{amsmath}
\usepackage{amssymb}
\usepackage{graphicx}
\usepackage{hhline}
\usepackage{multirow}

\newcommand{\methodname}{{Transfer learning-based Autoencoder with Residuals}}

\usepackage[colorlinks=true,linkcolor=blue,citecolor=blue]{hyperref}
\usepackage{color}
\usepackage{xcolor,colortbl}
\usepackage{soul}
\usepackage{enumitem}

\usepackage{multirow, float}
\usepackage[normalem]{ulem}
\usepackage[ruled,vlined]{algorithm2e}
\usepackage{comment}
\usepackage{booktabs}
\usepackage{array}
\newcolumntype{L}[1]{>{\raggedright\let\newline\\\arraybackslash\hspace{0pt}}m{#1}}
\newcolumntype{C}[1]{>{\centering\let\newline\\\arraybackslash\hspace{0pt}}m{#1}}
\newcolumntype{R}[1]{>{\raggedleft\let\newline\\\arraybackslash\hspace{0pt}}m{#1}}
\usepackage{lineno}

\begin{document}
%
\title{TAR: Generalized Forensic Framework to Detect Deepfakes using Weakly Supervised Learning}
%
\titlerunning{TAR: Generalized Forensic Framework to Detect Deepfakes}
%

 \author{Sangyup Lee
 \and
  Shahroz Tariq \and
 Junyaup Kim \and
 Simon S. Woo
 }
 \authorrunning{Sangyup Lee, Shahroz Tariq, Junyaup Kim, and Simon S. Woo}
 \institute{Sungkyunkwan University, Suwon, South Korea \\
 \email{\{sangyup.lee,shahroz,yaup21c,swoo\}@g.skku.edu}}

\maketitle              
\begin{abstract}
Deepfakes have become a critical social problem, and detecting them is of utmost importance. Also, deepfake generation methods are advancing, and it is becoming harder to detect. While many deepfake detection models can detect different types of deepfakes separately, they perform poorly on generalizing the detection performance over multiple types of deepfake. This motivates us to develop a generalized model to detect different types of deepfakes.
Therefore, in this work, we introduce a practical digital forensic tool to detect different types of deepfakes simultaneously and propose~\methodname~(TAR).
The ultimate goal of our work is to develop a unified model to detect various types of deepfake videos with high accuracy, with only a small number of training samples that can work well in real-world settings. 
We develop an autoencoder-based detection model with Residual blocks and sequentially perform transfer learning to detect different types of deepfakes simultaneously. Our approach achieves a much higher generalized detection performance than the state-of-the-art methods on the FaceForensics++ dataset. In addition, we evaluate our model on 200 real-world Deepfake-in-the-Wild (DW) videos of 50 celebrities available on the Internet and achieve 89.49\% zero-shot accuracy, which is significantly higher than the best baseline model (gaining 10.77\%), demonstrating and validating the practicability of our approach.
\end{abstract}
\section{Introduction}
\label{sec:intro}

Deepfakes generated by deep learning approaches such as Generative Adversarial Networks (GANs) and Variational Autoencoders (VAEs) have sparked initial amusement and surprise in the field. However, the excitement has soon faded away and worries arose due to potential misuse of these technologies. Such concerns have become real, and deepfakes are now popularly used to generate fake videos, especially pornography, containing celebrities' faces~\cite{news1,news2,tariq2021i}. In particular, these types of deepfakes are causing severe damage in the US, UK, and South Korea 
with 41\% of the victims being American and British actresses, and 25\% being female K-pop stars~\cite{news4}. These images and videos are starting to become widespread on the Internet and are rapidly propagating through social medias~\cite{news1,news3}. Therefore, deepfakes-generated videos raise not only significant social issues, but also ethical and privacy concerns. A recent report~\cite{study1} shows that nearly 96\% of deepfake videos are porn with over 134 million views.
However, despite the urgency of the problem, concrete media forensic tools for an effective detection of different types of deepfakes are still absent. Recently, several research methods have been proposed to detect GAN-generated images~\cite{Shahroz1,Shahroz2,SAM_ASOC} and deepfake  videos~\cite{guera2018deepfake,ciftci2020fakecatcher,tariq2020convolutional,tariq2021web,tariq2021i,nguyen2019capsule}. FaceForensic++~\cite{FaceForensics++} offers fake video datasets. 
Most of the detection methods generally perform well only in detecting fake images generated in the same domain (i.e., training and testing data are created with the same method), but not in detecting those generated in a different domain (i.e., test deepfakes are generated with different methods). So if new deepfake generation methods are developed, can the existing detection methods effectively detect the new deepfakes? How can we detect new fake videos? Is there a systematic and more generalizable approach to detect new types of deepfakes? The main focus of our work is to develop a generalized detection model, which not only performs well in detecting fake videos generated in one source domain, but also in detecting fake videos created in other domains with high accuracy. 

To achieve this, we propose a Transfer learning-based Autoencoder with Residuals (TAR) to improve the detection performance of different types of deepfakes. We also introduce a latent space Facilitator module with shortcut residual connections in the network to learn deep features of deepfakes effectively and apply transfer learning between different domains. Our few-shot sequential transfer learning procedure utilizes only a few data samples (only 50 frames, which is around 2 seconds of a video) to detect deepfake videos in different domains (e.g., transfer learning to Face2Face with FaceSwap trained model). 
Our contributions are summarized as follows:
\begin{itemize}
    \item We propose a~\methodname~(TAR) to improve the detection performance of three different types of deepfakes,
    achieving over 99\% accuracy and
    outperforming other baseline models.
    \item We sequentially apply transfer learning from one dataset to another to detect deepfake videos generated from different methods and achieve an average detection accuracy of 98.01\% with a single model for all deepfake domains, demonstrating superior generalizability over the state-of-the-art approaches. 
    \item Further, we evaluate and compare our approach using 200 real-world Deepfake-in-the-Wild videos collected from the Internet, and achieve 89.49\% detection accuracy, which is significantly higher than the best baseline models.
\end{itemize}

\begin{figure}[t]
\centering
\includegraphics[width=1\linewidth]{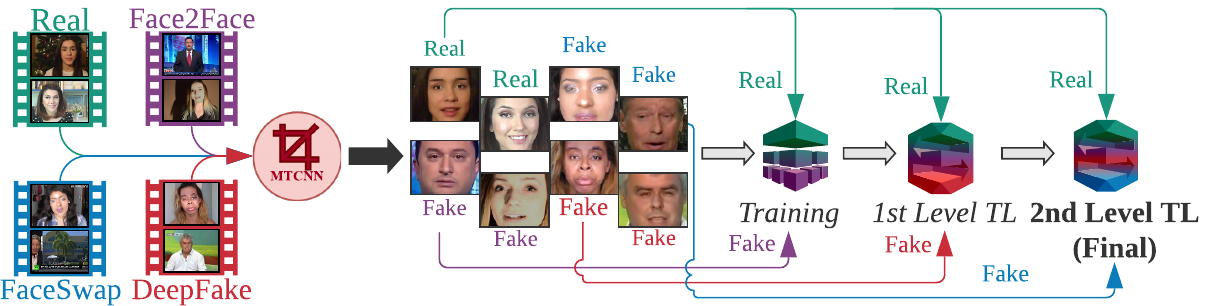}
\caption{Overview of our approach. Our proposed method learns features of deepfakes via Autoencoder with Residual blocks. Then, our multi-level transfer learning is sequentially performed with a small amount of training set in new target domains and the final model is created to effectively detect various types of deepfakes, including unseen or new deepfakes. }
\label{fig:overview}
\end{figure}
\section{Related Work}
\label{sec:related}
R\"ossler et al.~\cite{FaceForensics++} originally developed FaceForensics++, with 5,000 videos (1,000 Real and 4,000 Fake) using four different deepfake generation methods to benchmark deepfakes detection performance. 
Facebook also released Deep Fake Detection Challenge (DFDC) dataset~\cite{DFDC} with 5,250 videos (1,131 Real and 4,119 Fake), while Google released 3,363 videos (363 Real and 3,000 Fake) as part of the FaceForensics++ dataset~\cite{FaceForensics++}. 
For the detection of such deepfakes, deep learning-based methods in a supervised setting have shown high detection accuracy. Many Convolutional Neural Network (CNN)-based methods rely on content suppression using high-pass filters, either as a first layer~\cite{FT33} or using some hand-crafted~\cite{FT11} and learned features~\cite{FT7}. Also, image splice or visual artifacts detection methods~\cite{Splice1,FT36,matern2019exploiting} try to exploit the inconsistency arising from splicing near the boundaries of manipulated sections in the image. 
Zhou \textit{et al.}~\cite{FT44} built upon the idea of a high-pass filter method and attached a CNN-based model to capture both high-level and low-level details of the image. Tariq \textit{et al.}~\cite{Shahroz1,Shahroz2} proposed ShallowNet, an efficient CNN-based network to detect GAN-generated images with high accuracy even at a resolution as low as 64$\times$64. R\"ossler \textit{et al.}~\cite{FaceForensics++} significantly improved the performance on compressed images, which is essential for detecting deepfakes on social networking sites such as Facebook, Twitter, and WhatsApp. 
However, in the case of detecting new unseen deepfake generation methods, most approaches have failed to achieve high performance. 

To address this challenge, we explore few-shot transfer learning through our TAR model.
Since there will be new deepfake generation methods in the future, it would not be always feasible to collect and generate a large amount of new deepfake samples. To cope with this challenge, transfer learning (TL) with few-shot is the key to detect deepfakes generated through different methods (domains), that is, what has been learned in one domain (e.g., FaceSwap (FS)) can be exploited to improve generalization in another domain (e.g., Face2Face (F2F)). In this example sequence, we assume that Face2Face surfaced as a new deepfake method, with only a few videos to train. This approach enables the model to adapt to the new type of deepfake quickly.
Nguyen \textit{et al.}~\cite{nguyen2019multi} implemented fine-tuning with 140 videos to the pretrained model to detect unseen deepfake generation method. However, it is hard to construct a dataset including 140 videos for the newly emerging deepfake generation method in a realistic scenario.
Also, Cozzolino \textit{et al.}~\cite{ForensicTransfer} have explored the possibility of generalizing a single detection method to detect multiple deepfake target domains. 
In this work, we compare our approach against ForensicsTransfer~\cite{ForensicTransfer} to demonstrate an improved generalizability. In particular, we further compare our results with other approaches and test with unseen real-world Deepfake-in-the-Wild videos generated by unknown methods.

\section{Our Approach}
\label{sec:approach}
Our~\methodname~(TAR) model comprises a residual-based autoencoder with a latent space Facilitator module and transfers knowledge to learn deep features from different deepfake videos. The high-level overview of our approach is presented in Fig.~\ref{fig:overview}. 
We train our initial model on one source domain dataset, such as DeepFakes~\cite{Deepfakes} and apply transfer learning to the developed model using a small amount (few-shot learning) of data from another domain (e.g., FaceSwap~\cite{FaceSwap}) to effectively learn features from both domains. Similarly, we perform additional transfer learning with a small amount of data in other domains(e.g., Face2Face~\cite{Face2Face}). After a sequence of transfer learning processes, we obtain our final model, as shown in Fig.~\ref{fig:overview}, which can better detect deepfake videos generated by different methods (domains).

\subsection{Base Network Architecture}
We design our base network structure, as shown in Fig.~\ref{fig:arch}, and develop an autoencoder architecture consisting of Residual blocks with a Facilitator module to explicitly force and divide the latent space between real and fake embedding.
By comparing the activations in real and fake latent spaces, we can differentiate between real and fake images.

In addition, our proposed use of Residual blocks and Leaky ReLU is necessary to overcome the instabilities/weakness of prior approaches~\cite{Shahroz1,Shahroz2,ForensicTransfer}, such as ``zero activations'' (i.e., when both real and fake parts have no activation), while performing transfer learning experiments. 
We also discuss their limitations which motivate us to develop our TAR model in the Ablation Study section.
In particular, the TAR architecture is deeper, more stable, and is readily extendable with more shortcut connections to better capture deep features for different types of deepfake detection, compared to other approaches.
In our model, the encoder maps an image $\mathbf{x}$ to the latent space vector $\mathbf{H}$ and the decoder maps $\mathbf{H}$ to the reconstructed image $\mathbf{x'}$ as follows:
$\scriptstyle{\mathbf{H}=\sigma(\mathbf{W}\mathbf{x}+\mathbf{b})}$ and  $\scriptstyle{\mathbf{x'}=\sigma'(\mathbf{W'}\mathbf{H}+\mathbf{b'})}$, where $\mathbf{W}$ and $\mathbf{W'}$ are the weight matrices, $\mathbf{b}$ and $\mathbf{b'}$ are the bias vectors, and $\sigma$ and $\sigma'$ are the element-wise activation functions for the input images and the reconstructed images, respectively. Further, we divide the latent space $\mathbf{H}$ into two parts as follows: $\scriptstyle{\mathbf{H=\{H_1,H_2}}\}$, where we define the elements of the tensor $\mathbf{H_1}$ for real images and those of $\mathbf{H_2}$ for fake images at coordinates ($i$,$j$,$k$) by $\scriptstyle\mathbf{(H_1)}_{i,j,k}$ and $\scriptstyle\mathbf{(H_2)}_{i,j,k'}$. The dimension of each of the tensors $\scriptstyle{\mathbf{(H_1)}_{i,j,k}}$ and $\scriptstyle{\mathbf{(H_2)}_{i,j,k'}}$ is $\scriptstyle{M\times M \times N}$, and the dimension of $\scriptstyle{\mathbf{(H)}_{i,j,k}}$ is $\scriptstyle M \times M \times 2N$, where $\scriptstyle{1 \le i \le M,~1\le j \le M,~1\le k \le N}$, and $\scriptstyle{N+1\le k' \le 2N}$ (e.g., $\scriptstyle{M=15}$ and $\scriptstyle{N=64}$ are used in our work).

\subsubsection{Facilitator Module. }
For input image $\mathbf{x}_{m}$, the encoder generates the latent space output as follows: $\scriptstyle{Encoder(\mathbf{x}_{m}) = 
   \mathbf{H}_{m} =\{\mathbf{H}_{m,1}, \mathbf{H}_{m,2}\}.}$ In order to separate real and fake latent features, we implemented a Facilitator module shown in Fig.~\ref{fig:arch}, which forces the activations to be zeros for the opposite class and keeps the corresponding class. The activated $\scriptstyle{\mathbf{(H_m)}_{i,j,k}}$ by the Facilitator is defined as follows:  
\begin{equation}
    \small
    \label{eq:H1H2}
    (H_m)_{i,j,k} = 
    \begin{cases}
   (H_{m,1})_{i,j,k} & \text{ if } k \le N, c = 1~(real) \\
  (H_{m,2})_{i,j,k} & \text{ if } k > N, c = 2~(fake)\\
    0 & \text{ otherwise} \\   
    \end{cases},
\end{equation}
where $c$ is the class,  $\mathbf{(H_m)} \in {\rm I\!R}^{M \times M \times N}$ and $1\le i\le M, 1\le j \le M,$ and $1 \le k \le 2N$.

\subsubsection{Loss Functions. } We define two loss functions, $L_{Activ}$ for the real and fake activations, and $L_{Recon}$ for the reconstruction.
We combine these two loss functions with Residual blocks with skip connections to improve the performance:
\begin{equation}
\label{eq:loss}
    L = \lambda L_{Recon} + L_{Activ}.
\end{equation}
First, we define the per-class activation of the latent space $A_{m,c}$ 
to be the $L^1$ norm of $A_{m,c}/K_c$, where $K_c$ is the number of features of H$_{m,c}$ and $c \in$ \{1, 2\}. For the activation of an input $\mathbf{x}_{m}$ with a given label $l_m$ $\in$ \{1, 2\} and a latent feature tensor H$_{m,c}$, $L_{Activ}$ and $L_{Recon}$ are defined as follows:
\begin{equation}
\label{eq:act}
    L_{Activ} = \sum_{m}{\left |A_{m,2}-l_m+1\right |} + \sum_{m}{\left |A_{m,1}+l_m-2\right |},
\end{equation}
and we use $L^1$--norm for reconstruction loss ($L_{Recon}$),
\begin{equation}
\label{eq:recon}
     L_{Recon}=\frac{ \sum_{m}||{x_{m}-x_{m'}}||_{1}}{N}.
\end{equation}

\noindent
For an efficient embedding of $L_{Activ}$, we set $\lambda$ as 0.1 as determined empirically, thereby focusing more on embedding real and fake features through the encoder.

\subsubsection{Classification and Leaky ReLU. } 
\begin{figure}[t]
\centering
\includegraphics[clip, trim=0.0cm 0.12cm 0cm 0.05cm,width=1\linewidth]{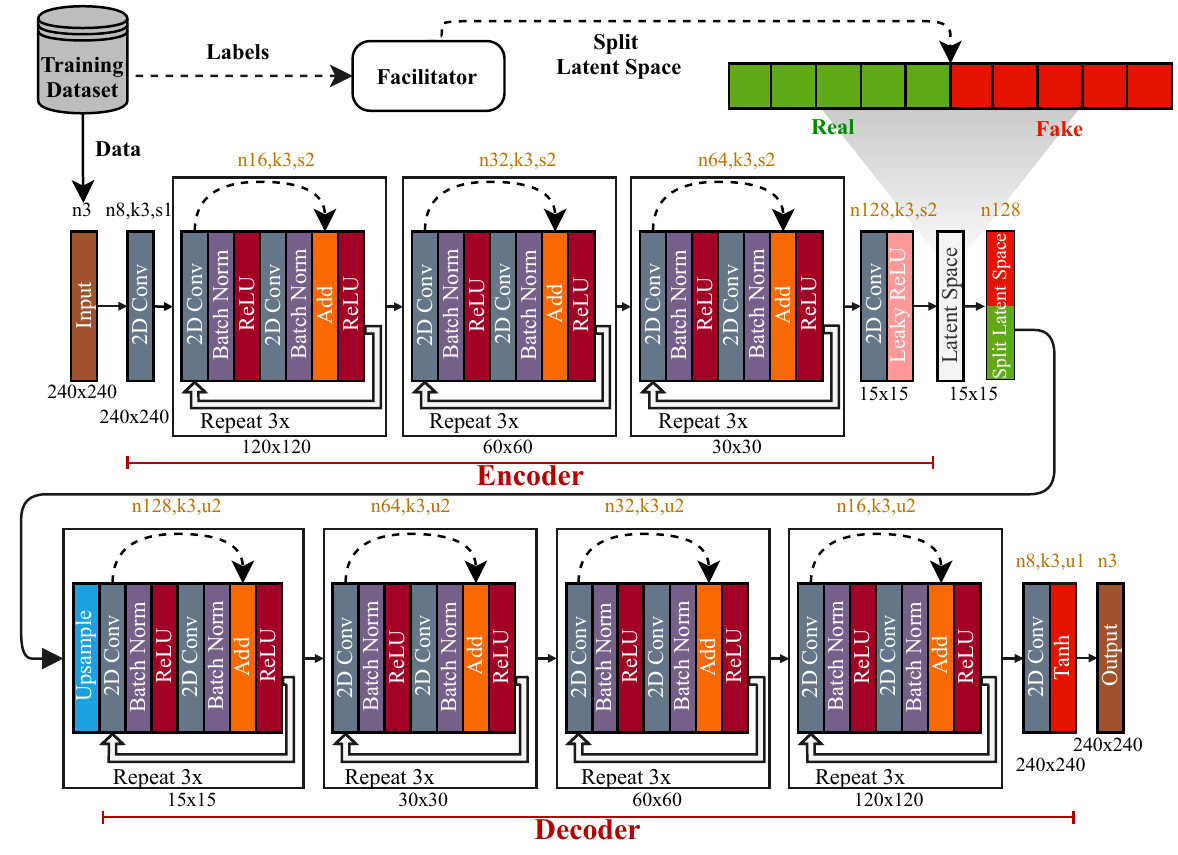}
\vspace{-10pt}
\caption{Our TAR model architecture. 
Depending on the label of the input, real or fake, different latent space representations are enforced by the Facilitator module.}
\label{fig:arch}
\end{figure}

In testing phase, if $A_{m,2}$ $>$ $A_{m,1}$, we consider the input $\mathbf{x}_{m}$ as fake. Otherwise, we classify it as real. However, for some small input images, we have empirically observed that $A_{m,1}$ = $A_{m,2}$ = 0 (zero activations). In this case, we cannot calculate the loss function or further perform classification. In order to prevent such situation, we use the Leaky ReLU function at the end of the last Convolution layer of the encoder and employ skip connections (see Fig.~\ref{fig:arch}), where the Leaky ReLU function with a small slope allows effectively calculating the loss and classifying very small input images.

\subsubsection{Residual Block with Shortcut Connections. }  
We implement Residual blocks with shortcut connections inside our autoencoder architecture, which can effectively learn deep features of images and keep the feature information through the network. As shown in Fig.~\ref{fig:arch}, we add three Residual blocks in the encoder and four Residual blocks in the decoder.
Each Residual block repeats three times and each block consists of a Convolutional (Conv) layer, a Batch Normalization (BN) layer, and a ReLU, which is followed by another Conv layer and BN layer. Next, we add the input to the Addition layer to build the skip connection and pass it through the final ReLU (see Fig.~\ref{fig:arch}).
After applying Residual blocks, the number of Conv layers increases to 45, which can lead to vanishing gradient problems. However, as mentioned by He et al.~\cite{resnet}, we can easily optimize a deeper network by adding the residuals or skip connections to the model structure.

\subsubsection{Encoder. } The input images have a resolution of 240$\times$240$\times$ 3 (see Fig.~\ref{fig:arch}). We use three Residual blocks (120$\times$120$\times$16, 60$\times$60$\times$32, and 30$\times$30$\times$64) in the encoder as shown in Fig.~\ref{fig:arch}, where each Residual block repeats three times.
The 2D Conv inside each Residual block, downsamples the input with a stride of 2 at the first loop (240$\times$240 to 120$\times$120, 120$\times$120 to 60$\times$60, etc.). After all the Residual blocks, we define one 3$\times$3 Conv layer followed by a Leaky ReLU with a small slope value of $10^{-7}$ to avoid the zero activation problem described earlier. At the end of the encoder, we chose $M$ = 15 and $2N$ = 128, to build a latent feature tensor ($\mathbf{H}$) with a size of 15$\times$15$\times$128. Therefore, the first 64 features ($\mathbf{H_1}$) are used to capture real images, and the other 64 features ($\mathbf{H_2}$), to represent fake images (see Fig.~\ref{fig:arch}). During training, for each input $\mathbf{x}_{m}$, the selection block is chosen by the Facilitator module according to the class label $l_m \in \{1,2\}$ to set zero values for non-activated class and activate only for the input class in Eq.~\ref{eq:H1H2}.

\subsubsection{Decoder.} We configured the decoder using Residual blocks such as the encoder in reverse order as shown in Fig.~\ref{fig:arch}. The difference from the encoder is that all Conv layers inside the decoder's Residual blocks have a stride of 1 since we are not reducing the size. To recover the original input image size, we scale up the input size by 2, using a 2$\times$2 nearest neighbors upsampling layer before the first Residual block. The kernel size of the Conv layer is set to be the same as that of the encoder network. In the last layer, a single 3$\times$3 Conv layer is used with $\tanh$ as the activation function.

\subsection{Multi-Level Sequential Transfer Learning} 
Since our goal is to develop an architecture that can detect deepfakes generated by different methods, we perform multi-level sequential transfer learning, as shown in Fig.~\ref{fig:overview}.
First, we train each model with respect to a specific generation method (domain) one at a time. For transfer learning, we performed few-shot learning from source domain $s$ to target domain $t$. We call it first-level transfer learning, denoted by  $s \to t$, where $s \neq t$, and $s,t\in\{FS, F2F, DF\}$. After this, we also conduct additional transfer learning toward a second target domain $u$ using the previous transfer learned model; we call it second-level transfer learning, which we denote it by ($s \to t)\to u$, where $s \neq t \neq u$, and $s,t,u\in\{FS, F2F, DF\}$. The $\to$ arrow denotes the domain transfer sequence, where we only use 50 fake frames (2 seconds of a video) from the target domain to perform sequential few-shot transfer learning.

The main objective of our work is to develop a deployable, unified deepfake detection framework. 
If a new deepfake generation method is surfaced, the amount of data for training the model will be limited. We overcome this scenario with our multi-level sequential few-shot transfer learning approach. The first and second-level transfer learning simulates this scenario where each level represents the situation when a new deepfake generation method is present. We only re-train on few frames from a new type of deepfake video and update the model, preserving all the previous detection capacity.
\section{Experiments}
\label{sec:exp}
We evaluated our method on a popular benchmark dataset to measure the performance of a \textit{single model} detecting various types of deepfakes simultaneously.

\subsection{Dataset}
\subsubsection{FaceForensics++ Dataset. } We utilized three different types of deepfake videos from Rössler et al.~\cite{FaceForensics++}.
For each type, we extracted a total number of 16,050 frames from the videos. The details of the dataset are provided in Table~\ref{tab:dataset_details}.

\subsubsection{Deepfake-in-the-Wild (DW) Dataset. } 
To validate the effectiveness and practicability of the detectors, we used 200 freely accessible Deepfake-in-the-Wild (DW) videos of 50 celebrities from the Internet. From the 200 DW videos, we extracted a total number of 8,164 frames.

\begin{table}[t]
\centering
\caption{Dataset summary for different experiments.
}
\label{tab:dataset_details}
\resizebox{1\linewidth}{!}{%
\begin{tabular}{l|c|c|c|c|c} 
\toprule
 \textbf{Category}  & \begin{tabular}[c]{@{}c@{}}\textbf{Total}\\\textbf{Videos} \end{tabular} & \begin{tabular}[c]{@{}c@{}}\textbf{Extracted}\\\textbf{Frames} \end{tabular} & \begin{tabular}[c]{@{}c@{}}\textbf{D1:~Base Dataset}\\ (Real:Fake) \end{tabular} & \begin{tabular}[c]{@{}c@{}}\textbf{D2:~TL Dataset}\\(Real:Fake) \end{tabular} & \begin{tabular}[c]{@{}c@{}}\textbf{D3:~Test Dataset}\\(Real:Fake) \end{tabular} \\ 
\hline
\textbf{Pristine (Real)}  & 1,000 & 16,050 & - & - & - \\
\textbf{DeepFake (DF)}  & 1,000 & 16,050 & 15,000:15,000 & 50:50 & 500:500 \\
\textbf{FaceSwap (FS)}  & 1,000 & 16,050 & 15,000:15,000 & 50:50 & 500:500 \\
\textbf{Face2Face (F2F)}  & 1,000 & 16,050 & 15,000:15,000 & 50:50 & 500:500 \\
\textbf{Deepfake-in-the-Wild (DW)}  & 200 & 8,164 & - & - & 8,164:8,164 \\
\bottomrule
\end{tabular}
}
\end{table}

\subsubsection{Dataset Settings. }
For training and testing, we divided the above-mentioned datasets into three categories (\textbf{D1}-\textbf{D3}) with no overlap to accurately characterize the performance. 
\begin{itemize}
    \item \textbf{D1: Base dataset.}
     The purpose of this dataset is to test the performance of a model for one source domain on which the model is originally trained. We also measure the zero-shot performance for the dataset on which the model is not trained.
     To evaluate the performance, we use 500 real and 500 fake images from the extracted frames as the test set, hence a total of three pairs of datasets, DF, FW, and F2F, as shown in Table~\ref{tab:dataset_details}. For simplicity, we refer these dataset pair with the fake class dataset name. For example, the Pristine and DeepFake dataset pair is referred to as DeepFake (DF).
     \item \textbf{D2: Transfer (few-shot) learning dataset.} We used the samples of 50 (2 seconds of a video) real and fake frames each. For evaluation, we used the test set from D3 (500 real and 500 fake frames) to check the model's performance.
     
     \item \textbf{D3: DW dataset.} We used this dataset only for testing the generalizability of our final model, as this dataset contains unseen videos generated with unknown deepfake generation methods. Therefore, this dataset can be served to assess realistic performance of the detection methods.
\end{itemize}

\subsection{Evaluation Settings}
\subsubsection{Baseline models. } We compared our method against ShallowNet~\cite{Shahroz2}, Xception~\cite{Xception}, and ForensicTransfer~\cite{ForensicTransfer}.

\subsubsection{Machine configurations. }
We used Intel i9-9900k CPU 3.60GHz with 64.0GB RAM and NVIDIA GeForce Titan RTX.
We tried our best to implement the baseline methods to their original description.

\subsubsection{Training and Multi-level Sequential Transfer Learning details. } We trained each model with respect to one specific generation method at a time. For transfer learning, we performed few-shot learning from source domain (DF or FS or F2F) to target domain (FS or F2F, or DF) as the first-level transfer learning and denote it as $s \to t$, where $s \neq t$, and $s,t\in\{FS, F2F, DF\}$. We also conducted transfer learning toward a second target domain using the previous transfer learned model as the second-level transfer learning. We denote second-level transfer learning as ($s \to t)\to u$, where $s \neq t \neq u$, and $s,t,u\in\{FS, F2F, DF\}$.

\section{Results}
\label{sec:results}
The detection performances of different methods are presented in Tables~\ref{tab:training_concise}-\ref{tab:TL2_concise}, where the highest accuracy values are in bold face.
The last `Avg. Gain' column in Table~\ref{tab:TL1_concise} and~\ref{tab:TL2_concise} reports the average accuracy gain from the best baseline model of each method in detecting all deepfake domain datasets.

\subsection{Base dataset performance (D1)} 
 We compared our approach with three baseline models using D1 as shown in Table~\ref{tab:training_concise}. First, after training with three source domains, our TAR model achieves higher accuracies of 99.80\%, 99.30\%, and 99.50\% in detecting DF, FS and F2F, respectively, compared to ShallowNet (63.00\%, 58.00\%, and 57.00\%), Xception (99.60\%, 92.00\%, and 99.30\%) and ForensicTransfer (FT) (99.80\%, 94.30\%, and 99.30\%). Generally, Xception and FT also performed well on base datasets, while ShallowNet performed poorly.

\begin{table}[t]
\centering
\caption{Accuracy of base (green) and zero-shot performance accuracy (gray).
}
\label{tab:training_concise}
\resizebox{1\linewidth}{!}{%
\begin{tabular}{L{2cm}|C{1cm}C{1.4cm}C{1.2cm}|C{1cm}C{1.4cm}C{1.2cm}|C{1cm}C{1.4cm}C{1.2cm}} 
\toprule
\multirow{2}{*}{\textbf{Methods} } & \multicolumn{3}{c|}{\textbf{Base Dataset: DF} (\%)} & \multicolumn{3}{c|}{\textbf{Base Dataset: FS} (\%)} & \multicolumn{3}{c}{\textbf{Base Dataset: F2F} (\%)} \\ 
\cline{2-10}
 & \textit{\textcolor[rgb]{0,0.502,0}{DF}}  & \textit{\textcolor[rgb]{0.502,0.502,0.502}{FS}}  & \textit{\textcolor[rgb]{0.502,0.502,0.502}{F2F}}  & \textit{\textcolor[rgb]{0,0.502,0}{FS}} & \textit{\textcolor[rgb]{0.502,0.502,0.502}{DF}} & \textit{\textcolor[rgb]{0.502,0.502,0.502}{F2F}}  & \textit{\textcolor[rgb]{0,0.502,0}{F2F}} & \textit{\textcolor[rgb]{0.502,0.502,0.502}{DF}} & \textit{\textcolor[rgb]{0.502,0.502,0.502}{FS}} \\ 
\hline
ShallowNet & \textcolor[rgb]{0,0.502,0}{63.0 } & \textcolor[rgb]{0.502,0.502,0.502}{48.0 } & \textcolor[rgb]{0.502,0.502,0.502}{53.0 } & \textcolor[rgb]{0,0.502,0}{58.0~} & \textcolor[rgb]{0.502,0.502,0.502}{44.0} & \textcolor[rgb]{0.502,0.502,0.502}{46.0 } & \textcolor[rgb]{0,0.502,0}{57.0} & \textcolor[rgb]{0.502,0.502,0.502}{51.0} & \textcolor[rgb]{0.502,0.502,0.502}{50.0} \\
Xception & \textcolor[rgb]{0,0.502,0}{99.6 } & \textcolor[rgb]{0.502,0.502,0.502}{50.0 } & \textcolor[rgb]{0.502,0.502,0.502}{50.3 } & \textcolor[rgb]{0,0.502,0}{92.0} & \textcolor[rgb]{0.502,0.502,0.502}{50.1} & \textcolor[rgb]{0.502,0.502,0.502}{51.5 } & \textcolor[rgb]{0,0.502,0}{99.3} & \textcolor[rgb]{0.502,0.502,0.502}{71.4} & \textcolor[rgb]{0.502,0.502,0.502}{50.4} \\
FT & \textbf{\textcolor[rgb]{0,0.502,0}{99.8}}  & \textcolor[rgb]{0.502,0.502,0.502}{50.0 } & \textcolor[rgb]{0.502,0.502,0.502}{70.4 } & \textcolor[rgb]{0,0.502,0}{94.3} & \textcolor[rgb]{0.502,0.502,0.502}{47.1} & \textcolor[rgb]{0.502,0.502,0.502}{53.7 } & \textcolor[rgb]{0,0.502,0}{99.3} & \textcolor[rgb]{0.502,0.502,0.502}{99.0} & \textbf{\textcolor[rgb]{0.502,0.502,0.502}{73.5}} \\
\textbf{TAR(\textit{Ours})} & \textbf{\textcolor[rgb]{0,0.502,0}{99.8}}  & \textbf{\textcolor[rgb]{0.502,0.502,0.502}{50.1}}  & \textbf{\textcolor[rgb]{0.502,0.502,0.502}{75.3}}  & \textbf{\textcolor[rgb]{0,0.502,0}{99.3}} & \textbf{\textcolor[rgb]{0.502,0.502,0.502}{50.8}} & \textbf{\textcolor[rgb]{0.502,0.502,0.502}{70.5}}  & \textbf{\textcolor[rgb]{0,0.502,0}{99.5}} & \textbf{\textcolor[rgb]{0.502,0.502,0.502}{99.7}} & \textcolor[rgb]{0.502,0.502,0.502}{52.2} \\
\bottomrule
\end{tabular}
}
\end{table}

\begin{table}
\begin{minipage}[b]{.485\textwidth}
   \centering
   \caption{Accuracy of first-level TL (blue) from base dataset (green).}
   
\begin{tabular}{|l|l|c|c|c|c|} 
\hline
\multicolumn{1}{|c|}{\multirow{2}{*}{\begin{tabular}[c]{@{}c@{}}\textbf{Transfer }\\\textbf{Seq.}\end{tabular}}} & \multicolumn{1}{c|}{\multirow{2}{*}{\begin{tabular}[c]{@{}c@{}}\textbf{Test}\\\textbf{Data}\end{tabular}}} & \multicolumn{3}{c|}{\textbf{Models}} & \multirow{2}{*}{\begin{tabular}[c]{@{}c@{}}\textbf{Avg. }\\\textbf{Gain}\end{tabular}} \\ 
\cline{3-5}
\multicolumn{1}{|c|}{} & \multicolumn{1}{c|}{} & Xce & FT & \textit{\textbf{TAR}} &  \\ 
\hline
\multirow{3}{*}{DF~$\to$~FS} & \textcolor[rgb]{0,0.502,0}{DF} & \textcolor[rgb]{0,0.502,0}{67.2} & \textcolor[rgb]{0,0.502,0}{90.1} & \textbf{\textcolor[rgb]{0,0.502,0}{91.6}} & \multirow{3}{*}{-2.0} \\ 
\cline{2-5}
 & \textcolor[rgb]{0,0,0.545}{FS} & \textcolor[rgb]{0,0,0.545}{54.0} & \textcolor[rgb]{0,0,0.545}{89.9} & \textbf{\textcolor[rgb]{0,0,0.545}{93.3}} & \\ 
\cline{2-5}
 & \textcolor[rgb]{0.502,0.502,0.502}{F2F} & \textcolor[rgb]{0.502,0.502,0.502}{62.6} & \textbf{\textcolor[rgb]{0.502,0.502,0.502}{90.9}} & \textcolor[rgb]{0.502,0.502,0.502}{80.1} & \\ 
\hline
\multirow{3}{*}{DF~$\to$~F2F} & \textcolor[rgb]{0,0.502,0}{DF} & \textcolor[rgb]{0,0.502,0}{95.8} & \textcolor[rgb]{0,0.502,0}{85.0} & \textbf{\textcolor[rgb]{0,0.502,0}{98.9}} &  \multirow{3}{*}{8.8}  \\ 
\cline{2-5}
 & \textcolor[rgb]{0,0,0.545}{F2F} & \textcolor[rgb]{0,0,0.545}{69.7} & \textcolor[rgb]{0,0,0.545}{81.6} & \textbf{\textcolor[rgb]{0,0,0.545}{93.1}} & \\ 
\cline{2-5}
 & \textcolor[rgb]{0.502,0.502,0.502}{FS} & \textcolor[rgb]{0.502,0.502,0.502}{53.5} & \textcolor[rgb]{0.502,0.502,0.502}{45.3} & \textbf{\textcolor[rgb]{0.502,0.502,0.502}{53.5}} & \\ 
\hline
\multirow{3}{*}{FS~$\to$~DF} & \textcolor[rgb]{0,0.502,0}{FS} & \textbf{\textcolor[rgb]{0,0.502,0}{92.0}} & \textcolor[rgb]{0,0.502,0}{48.2} & \textcolor[rgb]{0,0.502,0}{69.0} &  \multirow{3}{*}{17.5}  \\ 
\cline{2-5}
 & \textcolor[rgb]{0,0,0.545}{DF} & \textcolor[rgb]{0,0,0.545}{45.8} & \textcolor[rgb]{0,0,0.545}{79.8} & \textbf{\textcolor[rgb]{0,0,0.545}{98.5}} &  \\ 
\cline{2-5}
 & \textcolor[rgb]{0.502,0.502,0.502}{F2F} & \textcolor[rgb]{0.502,0.502,0.502}{68.0} & \textcolor[rgb]{0.502,0.502,0.502}{67.4} & \textbf{\textcolor[rgb]{0.502,0.502,0.502}{90.8}} & \\ 
\hline
\multirow{3}{*}{FS~$\to$~F2F} & \textcolor[rgb]{0,0.502,0}{FS} & \textbf{\textcolor[rgb]{0,0.502,0}{99.0}} & \textcolor[rgb]{0,0.502,0}{56.6} & \textcolor[rgb]{0,0.502,0}{84.6} &  \multirow{3}{*}{16.0}  \\ 
\cline{2-5}
 & \textcolor[rgb]{0,0,0.545}{F2F} & \textcolor[rgb]{0,0,0.545}{57.7} & \textcolor[rgb]{0,0,0.545}{55.1} & \textbf{\textcolor[rgb]{0,0,0.545}{84.7}} &  \\ 
\cline{2-5}
 & \textcolor[rgb]{0.502,0.502,0.502}{DF} & \textcolor[rgb]{0.502,0.502,0.502}{49.8} & \textcolor[rgb]{0.502,0.502,0.502}{56.5} & \textbf{\textcolor[rgb]{0.502,0.502,0.502}{85.3}} &  \\ 
\hline
\multirow{3}{*}{F2F~$\to$~DF} & \textcolor[rgb]{0,0.502,0}{F2F} & \textcolor[rgb]{0,0.502,0}{95.4} & \textcolor[rgb]{0,0.502,0}{98.7} & \textbf{\textcolor[rgb]{0,0.502,0}{99.2}} &  \multirow{3}{*}{1.7}  \\ 
\cline{2-5}
 & \textcolor[rgb]{0,0,0.545}{DF} & \textcolor[rgb]{0,0,0.545}{92.6} & \textcolor[rgb]{0,0,0.545}{99.0} & \textbf{\textcolor[rgb]{0,0,0.545}{99.5}} &  \\ 
\cline{2-5}
 & \textcolor[rgb]{0.502,0.502,0.502}{FS} & \textbf{\textcolor[rgb]{0.502,0.502,0.502}{68.7}} & \textcolor[rgb]{0.502,0.502,0.502}{60.9} & \textcolor[rgb]{0.502,0.502,0.502}{64.9} & \\ 
\hline
\multirow{3}{*}{F2F~$\to$~FS} & \textcolor[rgb]{0,0.502,0}{F2F} & \textcolor[rgb]{0,0.502,0}{87.6} & \textcolor[rgb]{0,0.502,0}{95.8} & \textbf{\textcolor[rgb]{0,0.502,0}{97.7}} &  \multirow{3}{*}{3.3} \\ 
\cline{2-5}
 & \textcolor[rgb]{0,0,0.545}{FS} & \textcolor[rgb]{0,0,0.545}{76.2} & \textcolor[rgb]{0,0,0.545}{92.7} & \textbf{\textcolor[rgb]{0,0,0.545}{97.4}} &  \\ 
\cline{2-5}
 & \textcolor[rgb]{0.502,0.502,0.502}{DF} & \textcolor[rgb]{0.502,0.502,0.502}{87.2} & \textcolor[rgb]{0.502,0.502,0.502}{94.4} & \textbf{\textcolor[rgb]{0.502,0.502,0.502}{97.8}} & \\
\hline
\end{tabular}
   \label{tab:TL1_concise}
\end{minipage}\quad
\begin{minipage}[b]{.485\textwidth}
   \centering
\caption{Accuracy of second-level TL (blue) from first-level (green).}
\begin{tabular}{|c|l|c|c|c|c|} 
\hline
\multirow{2}{*}{\begin{tabular}[c]{@{}c@{}}\textbf{Transfer}\\\textbf{Seq.}\end{tabular}} & \multicolumn{1}{c|}{\multirow{2}{*}{\begin{tabular}[c]{@{}c@{}}\textbf{Test}\\\textbf{Data}\end{tabular}}} & \multicolumn{3}{c|}{\textbf{Models}} & \multirow{2}{*}{\begin{tabular}[c]{@{}c@{}}\textbf{Avg. }\\\textbf{Gain}\end{tabular}} \\ 
\cline{3-5}
 & \multicolumn{1}{c|}{} & Xce & FT & \textbf{\textit{TAR}} &  \\ 
\hline
\multirow{3}{*}{\begin{tabular}[c]{@{}c@{}}(DF→FS)\\→F2F\end{tabular}} & \textcolor[rgb]{0,0.392,0}{DF} & \textcolor[rgb]{0,0.392,0}{88.2} & \textcolor[rgb]{0,0.392,0}{64.3} & \textbf{\textcolor[rgb]{0,0.392,0}{91.9}} & \multirow{3}{*}{17.1} \\ 
\cline{2-5}
 & \textcolor[rgb]{0,0.502,0}{FS} & \textcolor[rgb]{0,0.502,0}{55.1} & \textcolor[rgb]{0,0.502,0}{50.0} & \textbf{\textcolor[rgb]{0,0.502,0}{85.0}} &  \\ 
\cline{2-5}
 & \textcolor[rgb]{0,0,0.545}{F2F} & \textcolor[rgb]{0,0,0.545}{71.7} & \textcolor[rgb]{0,0,0.545}{62.3} & \textbf{\textcolor[rgb]{0,0,0.545}{89.4}} &  \\ 
\hline
\multirow{3}{*}{\begin{tabular}[c]{@{}c@{}}(DF→F2F)\\→FS\end{tabular}} & \textcolor[rgb]{0,0.392,0}{DF} & \textcolor[rgb]{0,0.392,0}{75.4} & \textcolor[rgb]{0,0.392,0}{90.0} & \textbf{\textcolor[rgb]{0,0.392,0}{97.8}} & \multirow{3}{*}{6.5} \\ 
\cline{2-5}
 & \textcolor[rgb]{0,0.502,0}{F2F} & \textcolor[rgb]{0,0.502,0}{55.0} & \textcolor[rgb]{0,0.502,0}{89.4} & \textbf{\textcolor[rgb]{0,0.502,0}{94.9}} &  \\ 
\cline{2-5}
 & \textcolor[rgb]{0,0,0.545}{FS} & \textcolor[rgb]{0,0,0.545}{67.5} & \textcolor[rgb]{0,0,0.545}{89.8} & \textbf{\textcolor[rgb]{0,0,0.545}{95.9}} &  \\ 
\hline
\multirow{3}{*}{\begin{tabular}[c]{@{}c@{}}(FS→DF)\\→F2F\end{tabular}} & \textcolor[rgb]{0,0.392,0}{FS} & \textcolor[rgb]{0,0.392,0}{48.8} & \textcolor[rgb]{0,0.392,0}{66.5} & \textbf{\textcolor[rgb]{0,0.392,0}{90.2}} & \multirow{3}{*}{18.1} \\ 
\cline{2-5}
 & \textcolor[rgb]{0,0.502,0}{DF} & \textbf{\textcolor[rgb]{0,0.502,0}{97.0}} & \textcolor[rgb]{0,0.502,0}{47.3} & \textcolor[rgb]{0,0.502,0}{84.9} &  \\ 
\cline{2-5}
 & \textcolor[rgb]{0,0,0.545}{F2F} & \textcolor[rgb]{0,0,0.545}{62.7} & \textcolor[rgb]{0,0,0.545}{57.8} & \textbf{\textcolor[rgb]{0,0,0.545}{87.8}} &  \\ 
\hline
\multirow{3}{*}{\begin{tabular}[c]{@{}c@{}}(FS→F2F)\\→DF\end{tabular}} & \textcolor[rgb]{0,0.392,0}{FS} & \textcolor[rgb]{0,0.392,0}{48.4} & \textcolor[rgb]{0,0.392,0}{77.5} & \textbf{\textcolor[rgb]{0,0.392,0}{96.8}} & \multirow{3}{*}{22.7} \\ 
\cline{2-5}
 & \textcolor[rgb]{0,0.502,0}{F2F} & \textbf{\textcolor[rgb]{0,0.502,0}{96.4}} & \textcolor[rgb]{0,0.502,0}{48.6} & \textcolor[rgb]{0,0.502,0}{85.2} &  \\ 
\cline{2-5}
 & \textcolor[rgb]{0,0,0.545}{DF} & \textcolor[rgb]{0,0,0.545}{64.7} & \textcolor[rgb]{0,0,0.545}{59.9} & \textbf{\textcolor[rgb]{0,0,0.545}{95.5}} &  \\ 
\hline
\multirow{3}{*}{\begin{tabular}[c]{@{}c@{}}(F2F→DF)\\→FS\end{tabular}} & \textcolor[rgb]{0,0.392,0}{F2F} & \textcolor[rgb]{0,0.392,0}{76.2} & \textcolor[rgb]{0,0.392,0}{94.0} & \textbf{\textcolor[rgb]{0,0.392,0}{98.5}} & \multirow{3}{*}{4.2} \\ 
\cline{2-5}
 & \textcolor[rgb]{0,0.502,0}{DF} & \textcolor[rgb]{0,0.502,0}{76.2} & \textcolor[rgb]{0,0.502,0}{93.6} & \textbf{\textcolor[rgb]{0,0.502,0}{97.4}} &  \\ 
\cline{2-5}
 & \textcolor[rgb]{0,0,0.545}{FS} & \textcolor[rgb]{0,0,0.545}{86.9} & \textcolor[rgb]{0,0,0.545}{93.9} & \textbf{\textcolor[rgb]{0,0,0.545}{98.3}} &  \\ 
\hline
\multirow{3}{*}{\begin{tabular}[c]{@{}c@{}}(F2F→FS)\\→DF\end{tabular}} & \textcolor[rgb]{0,0.392,0}{F2F} & \textcolor[rgb]{0,0.392,0}{92.7} & \textbf{\textcolor[rgb]{0,0.392,0}{97.4}} & \textcolor[rgb]{0,0.392,0}{97.3} & \multirow{3}{*}{8.4} \\ 
\cline{2-5}
 & \textcolor[rgb]{0,0.502,0}{FS} & \textcolor[rgb]{0,0.502,0}{66.6} & \textcolor[rgb]{0,0.502,0}{70.7} & \textbf{\textcolor[rgb]{0,0.502,0}{96.7}} &  \\ 
\cline{2-5}
 & \textcolor[rgb]{0,0,0.545}{DF} & \textcolor[rgb]{0,0,0.545}{95.9} & \textbf{\textcolor[rgb]{0,0,0.545}{97.8}} & \textcolor[rgb]{0,0,0.545}{97.1} &  \\
\hline
\end{tabular}
   \label{tab:TL2_concise}
\end{minipage}
\end{table}

\subsection{Zero-shot performance (D1)} Next, we also evaluated the base dataset-trained model with other domain datasets (i.e., testing the DF-trained model on F2F) for zero-shot performance comparison. The FT model achieved 73.50\% accuracy, which is 21.30\% higher than that of TAR for F2F-trained model tested on FS. However, in all other zero-shot testing cases, our TAR model outperformed all other approaches for all datasets as shown in Table~\ref{tab:training_concise}. Overall, our approach performed better than the-state-of-the-art methods in the same source domain, as well as different domains (zero-shot) except for one case. 

\subsection{Transfer learning performance (D2)} As shown in the last column of Table~\ref{tab:TL1_concise}, for the first-level transfer learning, we observed that our TAR model gained average accuracy over the best performing baseline (Xception or FT) on all sequences ($s\to t$) except for only one case (DF$\to$FS) in which FT achieved 90.30\% average performance.
By comparing Table~\ref{tab:training_concise} and~\ref{tab:TL1_concise}, we observed that after transfer learning to a target domain, the accuracy of the target domain increases, but the base class accuracy decreases for all methods. 
For example, when we trained all three models on the DF base dataset, all methods had a test accuracy of over 99\%, and a zero-shot accuracy of around 50\% on the FS dataset (see Table~\ref{tab:training_concise}).
Now, when we performed transfer learning from this source domain (DF) to target domain (FS), the accuracy of FT increases from 50.00\% to 89.90\%, and TAR increases from 50.10\% to 93.30\%. However, the source domain accuracy, which was 99.80\% for both FT and TAR, decreases to 90.10\% and 91.60\% respectively (see Table~\ref{tab:training_concise} and~\ref{tab:TL1_concise}). For Xception, the drop was even worse, with a 99.60\% to 67.20\% drop on the source domain, while gaining only 6\% on target domain. 
Compared to FT and TAR, Xception shows limited learning capacity during few-shot sequential transfer learning.

As shown in Table~\ref{tab:TL2_concise}, after the second-level transfer learning, our approach achieves a higher performance over every sequence (($s \to t$)$\to u$) compared to Xception and FT.
The best performing sequence for both TAR and FT has ``F2F$\to$DF'' as the source domain and FS as target domain, while Xception's best sequence is ``F2F$\to$FS'' as the source and DF as the target. We can write the resultant domain transfer sequence as ``(F2F$\to$DF)$\to$FS'' for TAR and FT, and ``(F2F$\to$FS)$\to$DF'' for Xception. The best performing FT model ``(F2F$\to$DF)$\to$FS'' achieved an average accuracy of 93.83\% on all three datasets, while the best performing TAR model ``(F2F$\to$DF)$\to$FS'' achieved 98.01\% gaining 4.2\% (see Table~\ref{tab:TL2_concise}).
However, Xception was not able to produce a similar generalization performance as FT or TAR, achieving an average accuracy of 85.01\%. Although there are 4 cases out of 18 in which the performances of Xception and FT are better than that of TAR, the overall performance of TAR is better. Therefore, our approach generalizes much better and yields above 97\% detection accuracy across all three datasets and shows significantly improved performance compared to zero-shot learning, as shown in Table~\ref{tab:training_concise}. Furthermore, FT's original work~\cite{ForensicTransfer} conducted experiments on multiple domains only on one case, whereas our research dealt with more complex cases to explore correlations between deepfake domains and to develop more generalized models.

\begin{table}[t]
\centering
\caption{Zero-shot performance on Deepfake-in-the-Wild (DW) dataset. The $\to$ arrow denotes the domain transfer sequence.
}
\label{tab:Pdataset_results}
\resizebox{1\linewidth}{!}{%
\begin{tabular}{l|c|c|c|c|c} 
\hline
 \textbf{Methods}             & \begin{tabular}[c]{@{}c@{}}\textbf{Domain Transfer}\\\textbf{Sequence~}(Best Performer) \end{tabular} & \begin{tabular}[c]{@{}c@{}}\textbf{Original}\\\textbf{Acc.} (\%)~~\end{tabular} & \begin{tabular}[c]{@{}c@{}}\textbf{Contrast (C)}\\\textbf{Acc.} (\%)\end{tabular} & \begin{tabular}[c]{@{}c@{}}\textbf{Brightness (B)}\\ \textbf{Acc.} (\%)\end{tabular} & \begin{tabular}[c]{@{}c@{}}\textbf{C + B}\\ \textbf{Acc.} (\%)~ ~\end{tabular}  \\ 
\hline
Xception                      & F2F$\to$FS$\to$DF                                                                                     & 50.23                                                                           & 58.11                                                                             & 66.92                                                                                & 78.72                                                                           \\
FT                            & F2F$\to$DF$\to$FS                                                                                     & 47.05                                                                           & 46.37                                                                             & 47.60                                                                                & 48.04                                                                           \\
\textbf{TAR (\textit{Ours})} & F2F$\to$DF$\to$FS                                                                                     & \textbf{67.87}                                                                  & \textbf{69.02}                                                                    & \textbf{77.06}                                                                       & \textbf{89.49}                                                                  \\
\hline
\end{tabular}
}
\end{table}

\subsection{Unseen DW dataset performance (D3)} It is critical to test the models against real-world deepfake data in addition to benchmark datasets.
Therefore, we used 200 Deepfake-in-the-Wild (DW) videos to evaluate the models in real-world scenarios. The model zero-shot detection accuracy on this dataset is shown in Table~\ref{tab:Pdataset_results}, where we used the best performing multi-level transfer learning sequence, ``(F2F$\to$DF)$\to$FS'', for both FT and TAR models and ``(F2F$\to$FS)$\to$DF'' for Xception.
As shown in the third column of Table~\ref{tab:Pdataset_results}, the accuracy of the second best performer from Table~\ref{tab:TL2_concise}, FT, was only 47.05\%. While TAR achieves an accuracy of 67.87\% and 50.23\% for Xception, which is higher than that of FT, it is much lower than the results for the previous datasets. 

Through our evaluation, this low performance strongly indicated that deepfakes in the wild were generated using more diverse, different, and complex methods than DF, F2F, and FS. Possibly, they could have leveraged additional post-processing methods to improve the quality of deepfakes. For example, we first noticed that the lighting in the DW dataset tends to be darker than that in the benchmark dataset. Hence, we first tried to balance the brightness and contrast of the DW videos~\cite{shorten2019survey} by increasing both the darkness and contrast by 30\% to better match the lighting conditions with the benchmark datasets. As shown in Table~\ref{tab:Pdataset_results}, increasing only the contrast yielded a slight accuracy improvement from 67.87\% to 69.02\% for the TAR model and 50.23\% to 58.11\% for Xception, but FT model has a drop in accuracy from 47.05\% to 46.37\%. After adjusting the brightness of the DW frames, our TAR model achieved a 77.06\% accuracy, which is almost 10\% higher than in the previous case. After applying both contrast and brightness changes together, Xception yields 78.72\% accuracy, while our approach achieves 89.49\%, which is much higher. On the other hand, the best performing baseline FT was not significantly improved, only achieving 48.04\% accuracy at best. Therefore, our approach adapting to new domains with multi-level transfer learning clearly outperforms other methods on unseen real-world dataset, demonstrating the effectiveness of TAR against real-world deepfakes. 

We plan to train our model with both brightness and contrast adjustments using data augmentation in the future. In this way, we can improve the performance of detecting DW or other new types of deepfakes without any inference-stage adjustments.

\section{Analysis \& Discussion}
\subsection{Classification Activation Map (CAM)} 
\begin{figure}[t]
\centering
\includegraphics[width=1\columnwidth]{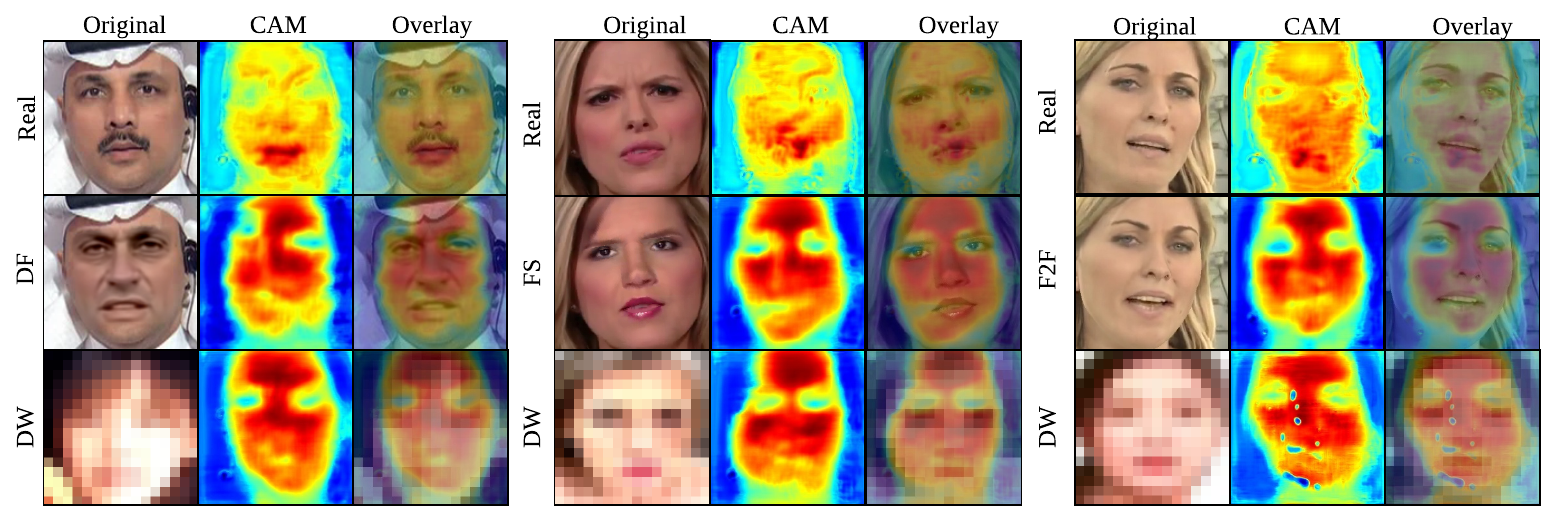}
\vspace{-15pt}
\caption{
A side-by-side comparison of 1) real vs. fake frames, 2) Class Activation Map (CAM) outputs, and 3) overlaid images of original input and CAM, using three different examples from each base dataset (DF, F2F, and FS). We also present sample images from the DW dataset in the same order, where we do not have an original real input image. The DW face images are intentionally blurred to hide the identity.
}
\label{fig:class_activation_map}
\end{figure}

To observe the fake frame activations of our model, we utilized CAM to generate the implicit attention of CNNs on an image. We used the best performing second-level transfer learned TAR model,
``(F2F$\to$DF)$\to$FS'', to show the activations from the decoder. In Fig.~\ref{fig:class_activation_map}, we present a side-by-side comparison of 1) original real versus fake images, 2) CAM outputs, and 3) overlaid images of original images and CAM, using three different examples from each dataset (DF, F2F, and FS). We also present sample images from the DW dataset in the same order, where we do not have an original real input image. The DW face images are intentionally blurred to hide the identity.
Figure.~\ref{fig:class_activation_map} clearly shows that our model produces much intense fake activation around the face than real images; the activated part is focused on the nose, the forehead, the cheek, etc. 
Since our model is a multi-level transfer-learned across DF, F2F, and FS, we did not observe any particular activation patterns for a specific model, and our approach detects all deepfake types effectively in a similar way.
\subsection{Ablation Study} We conducted an ablation study to check the impact of Leaky ReLU and residual connections in our model and summarized our finding. First, we removed the residual addition from our Residual blocks and replaced Leaky ReLU with ReLU. Our model started to show several zero activations in the last layer of the encoder for both real and fake images during training. We observed that this phenomenon occurred more frequently on small images.
Further research is needed to find the exact causes of this phenomenon. 
Next, we replaced the ReLU function with a Leaky ReLU and re-ran the same experiment. This time, there were much less zero activations, because the Leaky ReLU allows small negative values, instead of zeros. To completely eliminate this zero activation issue, we added back the residual addition inside Residual blocks. As a result, we did not observe any zero activation cases during training anymore, even with small, upscaled images. This ablation experiment supports our motivation to use Residual blocks and Leaky ReLU function for our TAR model.

\subsection{Discussion}
\label{sec:discussion}
Recently, deepfake porn videos have surged, with the vast majority featuring female celebrities' face photos to develop sexually explicit videos without the celebrities' knowledge or consent. These videos, rapidly spreading throughout the Internet, cause serious issues as they harass innocent people in the cyberspace. Because of the urgency of the problem and the absence of effective tools to detect these videos, we have undertaken this research to investigate real-world deepfake videos, in order to construct the Deepfake-in-the-Wild (DW) video dataset. We obtained freely available videos from the Internet, and all researchers in this study were informed about the detailed research protocol. We also consulted with the Institutional Review Board (IRB) in our institution, and received confirmation saying that approval was not required, since the videos are already available on the Internet. Moreover, we cropped the face regions and deleted the rest of the images, and further discarded the original videos. 

To preserve and protect the privacy of the celebrities, we blurred their faces in this work. Also, we do not plan on distributing any explicit content and leak celebrities' private information.

\subsection{Deployment}
To address the urgency of this matter, we have developed an online testing website since Dec. 2019 \footnote{\url{http://deepfakedetect.org}}, where people can upload images to check deepfakes using our model. We developed the interface such that it provides the probability of the image being fake and demonstrated that TAR can be practically deployed and used to circumvent the deployed real-world deepfakes.

\section{Conclusion}
\label{sec:conclusion}
Malicious applications of deepfakes, such as deepfake porn videos, are becoming increasingly prevalent these days.
In this work, we propose TAR to improve the generalized performance of multi-domain deepfake detection and test it on unseen real-world deepfake videos to evaluate its practicability.
A facilitator module splits our TAR model's latent space into real and fake latent spaces, enabling the encoder to focus more on learning the latent space representation, which results in a more accurate real and fake classification.
Our multi-level sequential transfer learning-based autoencoder with Residual blocks significantly outperforms other state-of-the-art methods in detecting multi-domain deepfakes from the FF++ dataset.
Furthermore, TAR achieves 89.49\% accuracy in detecting 200 real-world Deepfake-in-the-Wild videos of 50 celebrities, which is significantly higher than the state-of-the-art methods. 
Our results suggest that sequential learning-based models are an interesting venue to explore for analyzing and detecting fake and manipulated media in real-world settings and other analogous vision tasks that require domain adaptation. The code of our work is available on GitHub\footnote{\url{https://github.com/Clench/TAR_resAE}}.

\section*{Acknowledgments}
This work was partly supported by Institute of Information \& communications Technology Planning \& Evaluation (IITP) grant funded by the Korea government (MSIT) (No.2019-0-00421, AI Graduate School Support Program (Sungkyunkwan University)), (No. 2019-0-01343, Regional strategic industry convergence security core talent training business) and the Basic Science Research Program through National Research Foundation of Korea (NRF) grant funded by Korea government MSIT (No. 2020R1C1C1006004). Additionally, this research was partly supported by IITP grant funded by the Korea government MSIT (No. 2021-0-00017, Original Technology Development of Artificial Intelligence Industry) and was partly supported by the Korea government MSIT, under the High-Potential Individuals Global Training Program (2019-0-01579) supervised by the IITP.

\bibliographystyle{splncs04}
\bibliography{references}

\end{document}